\pdfoutput=1
\documentclass[11pt]{article}

\usepackage[final]{acl}
\usepackage{stfloats}
\usepackage{float}
\usepackage{times}
\usepackage{latexsym}
\usepackage{listings} 
\usepackage{tablefootnote}
\usepackage[T1]{fontenc}
\usepackage[utf8]{inputenc}

\usepackage{microtype}

\usepackage{inconsolata}

\usepackage{graphicx}

\title{SpeechLLMs for Large-scale Contextualized Zero-shot Slot Filling}

 \author{Kadri Hacioglu, Manjunath K E, Andreas Stolcke \\
          Uniphore   \\ \texttt{\{kadri.hacioglu, manjunath.ke, andreas.stolcke\}@uniphore.com} \\}

\begin{document}
\maketitle
\begin{abstract}
Slot filling is a crucial subtask in spoken language understanding (SLU), traditionally implemented as a cascade of speech recognition followed by one or more natural language understanding (NLU) components. The recent advent of speech-based large language models (speechLLMs), which integrate speech and textual foundation models, has opened new avenues for achieving speech understanding tasks in a more unified, generative, and instruction-following manner while promising data and compute efficiency with zero-shot abilities, generalizing to unseen slot labels. We address the slot-filling task by creating an empirical upper bound for the task, identifying performance, robustness, and generalization gaps, and proposing improvements to the training data, architecture, and training strategies to narrow the gap with the upper bound result. We show that each of these measures improve performance substantially, while highlighting practical challenges and providing empirical guidance and insights for harnessing these emerging models. 
\end{abstract}

\section{Introduction}
Modern conversational AI systems require a sophisticated integration of speech, language, and world knowledge combined with advanced understanding, reasoning, and generation capabilities to perform context-dependent NLP tasks effectively. This requires a tightly integrated model that can process long-form inputs empowered by high-quality representations of linguistic and world knowledge. 

Recently a unified approach that leverages both speech and text modalities, either in single-task or multitask configurations, has emerged, capitalizing on recent advances in text and speech foundation models \cite{brown2020languagemodelsfewshotlearners, touvron2023llamaopenefficientfoundation, openai2023gpt4, naveed2024comprehensiveoverviewlargelanguage, hsu2021hubert, 9814838, Mohamed2022} trained on web-scale datasets, which yield both data and computational efficiency. 

We ues the term speech large language model (speechLLM) for an architecture that tightly couples speech and text modalities through a decoder-only large language model, for speech recognition and understanding tasks generating textual outputs. This tight coupling addresses the challenges of error propagation, information loss and disjoint optimization in the traditional loosely coupled, cascaded systems.   

\begin{figure}[t]
  \centering
  \includegraphics[width=\linewidth]{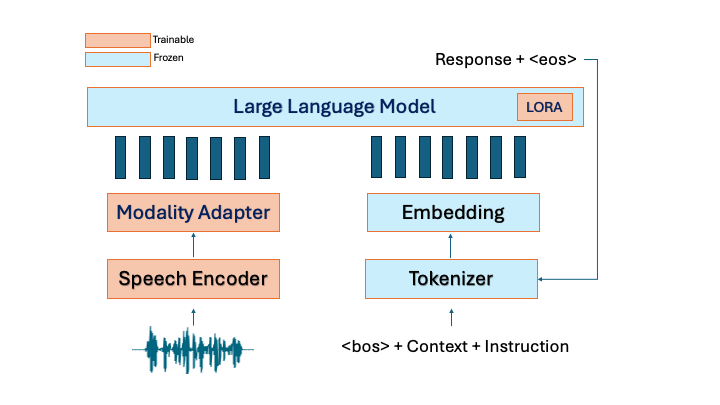}
  %\vskip -0.5cm
  \caption{SpeechLLM architecture}
  \label{fig:architecture}
\end{figure}

Slot filling is a key spoken language understanding (SLU) task in goal-oriented conversational systems where a model extracts structured information by identifying and extracting the slot labels and their values within the user's query~\cite{tur2011spoken}. This information is essential for allowing the system to understand user requests and intents to take appropriate actions toward fulfilling their goals. Traditionally, these systems, either in cascade or end-to-end form, were typically designed for specific closed domains for a predefined set of slot types, limiting their adaptability and generalization to new domains or even for evolving requirements within the same domain. 

Therefore, a model's zero-shot ability in slot filling, i.e., its capacity to generalize to unseen slot labels without having seen them in training, is crucial for deployment in dynamic environments where new user intents and slot types frequently emerge. SpeechLLMs are a natural solution for this task, allowing unseen labels to be specified via text modality and using instructions to capture their meaning and relationship to linguistic constructs during inference. 

In our work, among many possible speechLLM architectures \cite{peng2024surveyspeechlargelanguage, cui2025recentadvancesspeechlanguage, gong2023listen, tang2024salmonn, 10447112, 10447553, chu2024qwen2audiotechnicalreport, shu2023llasmlargelanguagespeech, 10389705, 10389703, hu-etal-2024-wavllm, 10832146}, we considered and implemented one that has a structure very similar to SpeechVerse \cite{das2024speechverselargescalegeneralizableaudio}.

As shown in Figure~\ref{fig:architecture}, it consists of three main components: a speech encoder, a modality adapter, and a pretrained large language model (LLM). This architecture is designed to jointly process speech signals and textual instructions, enabling it to perform a wide range of tasks that require understanding of both speech and text. The system combines the processed audio features from the pretrained speech encoder and modality adapter with the text embeddings along the time/position dimension to feed them into the LLM. This allows the model to generate an output probability distribution conditioned on both speech and text. The modality adapter transforms the speech modality into a representation compatible with the linguistic input space of the LLM. The foundational models, pretrained at scale in their respective modalities, promise data-efficient development. Moreover, small-scale modality adapters for alignment and large-scale foundational models with parameter-efficient fine-tuning enable computationally efficient development. Importantly, the emergent abilities of LLMs, if preserved during fine-tuning, offer the potential for instruction-based zero-shot learning and improved generalization capabilities, as well as general linguistic knowledge to interpret slot names. This approach not only addresses the data scarcity issue, but also enhances the adaptability of speech understanding systems to new domains and evolving requirements, marking a significant advancement in the field of spoken language understanding. This unified approach offers two key advantages: (1) simplified development and maintenance through a single integrated model architecture, and (2) improved data and compute efficiency that significantly reduce development costs.

In this paper, we first introduce a novel slot-filling dataset for the training of speechLLMs, supporting a large number of slot types at industrial scale for diverse call-center use cases and domains, while targeting zero-shot generalization to new labels, enabling faster time-to-delivery without specific training. Second, we present our novel prompting strategy to curate and annotate this dataset using LLMs, contextualized by inputting the complete call instead of individual turns. Then we present our systematic and stepwise approach of making architectural decisions, selecting training methods with various strategies to improve upon baseline performance, approaching the \textit{oracle} performance. Finally, we point to an important and overlooked distinction between \textit{composite} and \textit{foundational} speechLLMs and evaluate their relative performances based on fine-tuning of both models beyond considering just the out-of-the-box evaluations. Our experimental results show that
\begin{itemize}
    \item the complexity of modality adapters matters;
    \item modality alignment between multimodal input and response generation is crucial;
    \item multistage training is necessary for faster and better modal alignment
    \item extending the task-specific dataset to a carefully curated multitask dataset is essential for robustness and generalization.
    \item starting from \textit{foundational} speechLLMs result in a significant performance advantage, though with reduced flexibility.
\end{itemize}

\section{Data Preparation}
\subsection{Description and Annotation }

The data for slot filling is a set of scripted call-center conversations between agents and customers curated by DefinedAI\footnote{https://www.defined.ai}. This collection, denoted CallCenter-A, has approximately 31K calls with almost 1M turns and 2.1K hours of speech. The domains covered are banking, telecommunication, insurance and retail. We used GPT-4o to annotate our dataset with slot labels and values. To achieve broad, diverse and open-ended slot filling we decided not to prime the LLM for any specific set of slot labels. Instead, given the entire call itself, we instructed the LLM to do slot filling turn-by-turn for mentions that reflect real world entities, events, dates, times and  numerals while avoiding abstract notions of “entities” such as issues, solutions, broader concepts, advice or ideas. An example LLM prompt for this annotation is given in Appendix~\ref{sec:appendixB}.
\begin{table*}[t]
    \centering
    %\small
    \begin{tabular}{lccc}  % l=left, c=center for three columns
        \hline
        \textbf{Task} & \textbf{Dataset} & \textbf{Duration} & \textbf{Samples} \\
        \hline
        Slot Filling & CallCenter-A &  400 & 454  \\
        AST  & Internal & 400 & 172\\
        SIT & Spoken Alpaca & 30 & 29 \\
        SQIT & Spoken Alpaca & 14 & 14 \\
        \hline
        Total & All &  844 & 454 \\
        \hline
    \end{tabular}
     \caption{Training datasets: The audio durations are in hours and the number of training samples are in thousands.}
    \label{tab:training_dataset}
%\vspace{-15pt}
\end{table*}
\subsection{Instruction-based Training Dataset}
We now describe the conversion of the slot filling dataset into an instruction-based training dataset. This dataset consists of three fields: audio, instruction, and desired output. Instructions are the most crucial part of the data preparation. It should consist of a description of the task in natural language. Although it is possible to use a fixed instruction, a diverse set of instructions is beneficial for generalization to prompts unseen during training. In addition, we introduced several strategies for improving coverage across a variety of use cases of slot filling. We assume turn-by-turn slot filling with/without context, and with/without querying specific slots. The context is defined as the recognition results for the previous $T$ turns. We randomize the context size in the range $0 \le T \le 3$. When we query specific slots in the prompt we take the ground truth slots in the corresponding turn, if any, and then add a varying numbers $S$ of distractors. We also randomize the number of distractors, $1 \le S \le 5$. For each case, we randomly sample from a set of 10 prompts. Some examples of prompts are given in Appendix~\ref{sec:appendixB}.

% insert an example of (input text, instruction, output here) -- put in appendix is needed

\subsection{Multitask  Datasets}
In addition to the slot-filling dataset we considered other datasets for the following additional tasks: automatic speech transcription (AST), spoken instruction task (SIT), and spoken query instruction task (SQIT). %The AST dataset is a small subset  of our internal English dataset.
The "internal AST dataset" is a subset of our English training corpus for production models, chosen to represent the acoustic quality and language characteristics of call-center domains, disjoint from the slot-filling data. The other datasets are spoken versions of textual Alpaca instruction datasets converted by a text-to-speech system and open-sourced by \citet{li2024whismaspeechllmperformzeroshot}. These two datasets are described in detail in Appendix~\ref{sec:appendixA2}.
The main task is slot filling, with other datasets as auxiliary tasks to facilitate modality alignment and to prevent overfitting. Table \ref{tab:training_dataset} summarizes the training datasets used in our experiments.

\subsection{Evaluation Datasets}
In addition to the CallCenter-A evaluation dataset, we created another dataset from real call-center conversations to represent ``unseen'' acoustic and linguistic content, comprising 80 calls with a total of 4495 turns. It is annotated similarly to CallCenter-A using GPT-4o. While CallCenter-A is considered in-domain both acoustically and linguistically, CallCenter-B  differs in the acoustic conditions and contains slot labels that overlap only 48\% with CallCenter-A. Accordingly, we split CallCenter-B into two subsets: in-domain (CallCenter-B "ID") and out-of-domain (CallCenter-B "OOD"). Detailed information about these datasets is given in Appendix~\ref{sec:appendixA}.

\section{Experiments}

\subsection{Experimental Details}
Model development and experiments were run on four NVIDIA A10G GPUs (24GB each) using the Hugging Face ecosystem. To deal with limited resources, we employed parameter-efficient fine-tuning (PEFT) with QLoRA (4-bit quantization), supported by Accelerate and DeepSpeed. LoRA was configured with rank 32, $\alpha=128$, and dropout 0.05. Training used an effective batch size of 128 (batch 4 per GPU, accumulation 8), cosine learning rate scheduling ($2\times10^{-4}$ max, 20\% warm-up) over 10 to 15 epochs, and AdamW with default parameters. Gradient clipping was applied at 1.0.

Experimental configurations, settings and model selections are described in detail in Appendix~\ref{sec:appendixC}. 
The results presented in the following sections belong to the model with the best evaluation loss during fine-tuning over 10 to 15 epochs.

\begin{table*}[t]
    \centering
    %\small
    \begin{tabular}{lc}  % Adding vertical lines
        \hline
        \textbf{System} & \textbf{CallCenter-A Prec/Recall/F1}\\
        \hline
        Vanilla LLM & 0.0436/ 0.7259/ 0.0823  \\
        FT-LLM  & 0.8523/ 0.9901/ 0.9160 \\
        Whisper | FT-LLM & 0.7006/ 0.8982/ 0.7872\\
        SpeechLLM (baseline)  & 0.3892/ 0.4151/ 0.4017 \\
        \hline
    \end{tabular}
    \caption{Baseline system performance: Slot-filling accuracy given by partial-match precision/recall/F1 metrics. The fine-tuned Whisper operates at 14.20\% word error rate. FT-LLM denotes the fine-tuned LLM.}    
    \label{tab:baselines}
\vspace{-5pt}    
\end{table*}
%\vspace{-2pt}
\subsection{Baseline Systems and Performances}
We considered three baseline systems. The first is a textual slot-filling system using Llama 3.2 1B LLM. The model is fine-tuned using the textual part of CallCenter-A. The second baseline system is the traditional pipeline of ASR followed by an NLU component with
whisper-base~\cite{radford2022robustspeechrecognitionlargescale} model as the ASR component and Llama 3.2 1B for NLU. We used the internal AST dataset to fine-tune the Whisper-base model. The NLU component is the same textual LLM that was trained in the first baseline system. The third baseline system is a speechLLM using a simple modality adapter based on convolutional neural networks (CNNs). We employed a training strategy keeping the LLM and the speech encoder frozen and train only the adapter component using the dataset specific to slot filling. In addition, for better compute efficiency we employed 8x subsampling of the adapter output. 

The performance of the first (text-based) baseline system will serve as an upper bound on performance. Instead of ground truth transcripts, the LLM in the cascaded system uses speech recognition results, and the speechLLM uses adapted speech embeddings 

Table \ref{tab:baselines} shows the baseline results. The performance of the vanilla LLM is also included to highlight its inability to perform this highly structured task in a zero-shot manner. The speechLLM model was initially designed with a limited computational budget in mind, which explains its inferior performance. Ultimately, the speechLLM is required to generate the same response as the textual LLM but using the audio input, assuming successful modality alignment is achieved and the audio is ``heard'' correctly. In the following sections we will discuss, and experiment with, measures for narrowing the gap to the upper bound, while still staying within our computational constraints.

\subsection{Modality Adapters}
We investigated four adapters of increasing parametric complexity. The first adapter employs three 1-D convolutional layers (kernel=3, stride=2, channels=512) followed by a linear projector. The second consists of a speech padding and stacking component with a factor of 4, followed by a single linear projector. The third replaces the linear projector with a two-layer MLP incorporating a SwiGLU nonlinearity~\cite{shazeer2020gluvariantsimprovetransformer}. The fourth adapter is a 2-layer transformer with 8 attention heads. 

The parameter counts for the adapters are listed in Table~\ref{tab:adapter_params}. Rather than equalizing parameter counts, we designed each adapter to achieve a 8x subsampling rate while projecting the speech encoder output to match the LLM's input dimension. Each adapter processes temporal and frequency information differently: the CNN adapter captures local correlations while downsampling; the linear adapter models global correlations across sub-sampling segments linearly; the MLP adapter introduces nonlinear modeling of global correlations; and the transformer adapter uses self-attention mechanisms to capture both local and global dependencies while modeling complex temporal relationships through its multihead attention and feedforward networks.

The results in Table \ref{tab:adapter_params} suggests that the size of the adapter plays a crucial role in performance. To investigate the impact of the adapter type, we increased CNN size by adjusting the number of channels and kernel size to match the best-performing network. Despite these adjustments, the modified model, CNN-XL, exhibited relatively inferior performance, underscoring the importance of not only the model size, but also its design and modeling capability. 

\begin{table*}[tb]
    \centering
   % \small
    \begin{tabular}{lrc}
        \hline
        \textbf{Adapter} & \textbf{Parameters} & \textbf{Prec/Recall/F1}\\ 
        \hline
        CNN (baseline) & 3.41M &  0.3892/ 0.4151/ 0.4017 \\
        Linear & 8.39M & 0.5411/ 0.8396/ 0.6581\\
        MLP & 20.98M &  0.5758/ 0.8830/ 0.6970\\ 
        Transformer & 67.16M & 0.6214/ 0.9207/ 0.7420\\
        CNN-XL & 68.17M & 0.6232/ 0.7968/ 0.6994\\
        \hline
    \end{tabular}
    \caption{Performance and parameter count comparisons for different adapter architectures. Performance is measured on CallCenter-A data.}
    \label{tab:adapter_params}
\end{table*}

\subsection{LLM Fine-tuning}
Training an adapter with frozen foundational models targets only the alignment of the input modalities generating the desired output response for the downstream task. Since the LLM is frozen, its inherent ability to perform the downstream task determines the performance of the model. We hypothesize that, while aligning the modalities in their input representations, it is also crucial to align the input modalities with the desired LLM output. This requires the adaptation of the LLM. 

Table \ref{tab:adapter_lora} shows the results with LoRA adaptation enabled. It is noteworthy that the performance of all adapters increased significantly, except for the transformer adapter. Inspection of its learning curve showed divergent behavior that ended up at a local minimum with inferior performance. We believe that this is a natural consequence of the difficulty of joint training as the scale and complexity of the model increase. We elaborate on these training difficulties in Appendix~\ref{sec:appendixD}.

\begin{table*}[tb]
    \centering
    %\small
    \begin{tabular}{lccc}
        \hline 
        \textbf{Adapter Type} & \textbf{Prec/Recall/F1 with LoRA} & \textbf{$\Delta$F1}\\ 
         \hline
        CNN & 0.5304/ 0.8908/ 0.6649 & +0.2632 \\
        Linear  &  0.5761/ 0.9150/ 0.7071 & +0.0490\\
        MLP† &  0.6553/ 0.9268/ 0.7677 & +0.0707\\ 
        Transformer*‡ & 0.4272/ 0.7734/ 0.5504 & -0.1916 \\
        \hline
        \multicolumn{3}{p{0.9\linewidth}}{\small{* After the stabilization of learning by reducing the learning rate the performance is Precision: 0.6010, Recall: 0.9109, F1: 0.7242, $\Delta$F1: -0.0178  }} \\
        \multicolumn{3}{p{0.9\linewidth}}{\small{‡ After increasing the warm-up period the performance is Precision: 0.6228, Recall: 0.9357, F1: 0.7479  $\Delta$F1: 0.0059 }} \\
        \multicolumn{3}{p{0.9\linewidth}}{\small{† For the rest of our experiments with LoRA we continued with this best performing setup using MLP}}      
    \end{tabular}
        \caption{Performance comparisons across different adapter architectures when LoRA is enabled for LLM fine-tuning. Performance is reported on CallCenter-A data and the $\Delta$F1 is relative to the F1 values in Table~\ref{tab:adapter_params}.}
    \label{tab:adapter_lora}
%\vspace{-15pt}
\end{table*}
\begin{table*}[tb]

    \centering
    %\small
    \begin{tabular}{lc}  % Adding vertical lines
        \hline
        \textbf{Training strategy} & \textbf{Prec/Recall/F1}\\
        \hline
        Joint, single-stage &  0.6553/ 0.9268/ 0.7677 \\
        Multistage-A  & 0.6722/ 0.9297/ 0.7803 \\
        Multistage-B & 0.6653/ 0.9448/ 0.7808\\
        Multistage-C & 0.6842/ 0.9391/ 0.7916 \\
        \hline
    \end{tabular}
    \caption{CallCenter-A results for multistage training}    
    \label{tab:multi_stage_methods}
\end{table*}
\subsection{Multistage Training}
Several studies \cite{das2024speechverselargescalegeneralizableaudio, grattafiori2024llama3herdmodels} suggest training the models in a multistage manner for faster and more stable learning and better task performance. In this section we consider three strategies for multistage training. 

First, we demonstrate a multistage training strategy, namely Multistage-A, by (Stage1-a) fine-tuning the Whisper encoder, using (speech, text) pairs, and (Stage1-b) the LLM, using (instruction, response) pairs, followed by (Stage2) jointly fine-tuning both the modality adapter and the LLM using (speech, instruction, response) triplets. Note that the first-stage speech encoder and LLM actually correspond to the components used in the cascaded baseline system. In other words, the job of the first stage is to create fine-tuned foundational models separately and use them as the initial models for training the speechLLM in the second stage. 
\begin{table*}[b]
    \centering
     %\small
    \begin{tabular}{lcc}  
        \hline
        \textbf{System} & \textbf{CallCenter-B} & \textbf{CallCenter-B} \\
         & \textbf{ID} & \textbf{OOD} \\
        \hline
        FT-LLM & 0.8360/ 0.9530/ 0.8907  & 0.5439/ 0.8474/ 0.6626\\
        Whisper | FT-LLM  & 0.5909/ 0.7067/ 0.6436 & 0.4280/ 0.6278/ 0.5090\\
        SpeechLLM & 0.3964/ 0.7306/ 0.5140 & 0.2386/ 0.4500/ 0.3119  \\
        SpeechLLM++ & 0.4240/ 0.7810/ 0.5496 & 0.2645/ 0.5448/ 0.3561  \\
        \hline
    \end{tabular}
        \caption{Performance on out-of-domain datasets. The system marked with ``++'' use multitask data expansion described in Section~\ref{sec:data_expansion}. SpeechLLM corresponds to the singel-stage training strategy.}
    \label{tab:ood}
\end{table*}

Second, we consider alignment strategy Multistage-B, where we first (Stage1) fine-tune the adapter on a new task cast as a ``continuation'' task of the audio transcripts. This method is similar to the method proposed in~\citet{fathullah2024audiochatllamageneralpurposespeechabilities}. We present the transcripts to the foundational LLM as a prompt and run it to complete the transcripts in an auto-regressive manner creating (transcript, continuation) pairs. Then, we convert this into a fine-tuning dataset for the speechLLM by replacing transcripts with audio as (audio, continuation) pairs. Using the continuation outputs with audio to fine-tune the adapter ensures an alignment that both spoken and textual inputs to the LLM generate the same response. After this alignment, the adapter and LLM are jointly (Stage2) fine-tuned further to revise the aligned modality to generate task-specific outputs. 

Finally, we experiment with a third multistage strategy, Multistage-C, in which we first (Stage1) fine-tune the adapter using an AST task  to align the audio encoder and Llama 3.2 1B in the linguistic embedding space. This is followed by (Stage2) joint adaptation of adapter and the language model. 

Table~\ref{tab:multi_stage_methods} shows the results for all the methods, highlighting the importance of multistage strategies for improved performance compared to a single-stage strategy. In Appendix~\ref{sec:appendixE} we also illustrate through learning curves their faster convergence to a better operating point .

\subsection{Out-of-domain Performance}

We also ran experiments to probe the generalization, robustness and zero-shot capabilities of the models using three distinct datasets we briefly mentioned earlier and further described in detail in Appendix~\ref{sec:appendixA}. 

The results in Table~\ref{tab:ood} show that despite being trained on the same data and using the same components, the cascaded system remains more robust, generalizes better, and exhibits stronger zero-shot capabilities than the bimodal speechLLM architecture. This performance gap suggests that the bimodal model's scale, the size of training data and the training process are not yet sufficient for joint and tightly coupled integration of information. This highlights that modality alignment and LLM fine-tuning may not be sufficient for the joint model to perform better than the sequential performance of individual unimodal models, unless we employ training at scale (more tokens and parameters).  
\vspace{2pt}
\subsection{Multitask Training}
    \label{sec:data_expansion}
To enhance multimodal alignment without increasing the model size, we expanded the task-specific dataset with additional auxiliary tasks, following insights from related work by \citet{li2024whismaspeechllmperformzeroshot}. As described in Section 3.3, we added the AST task as a simpler task to strengthen cross-modal alignment, while spoken instruction/query tasks were added to mitigate overfitting in the text-based foundational model. This strategy aimed to balance the fine-tuning, ensuring better robustness and generalization under the constrained model capacity.  

The results clearly show that training with auxiliary tasks significantly improves performance on both in-domain and out-of-domain data sets. The results are included in Tables~\ref{tab:ood} and \ref{tab:data_expansion}.

\begin{table*}[h]
    \centering
    %\small
    \begin{tabular}{lc} 
        \hline
        \textbf{System} & \textbf{CallCenter-A Prec/Recall/F1}\\
        \hline
        SpeechLLM &  0.6553/ 0.9268/ 0.7677 \\
    
        SpeechLLM++ &  0.6632/ 0.9458/ 0.7797 \\
    \end{tabular}
  \vspace{-4pt}
    \caption{Performance with multitask training. ``++''denotes fine-tuning with multitask data expansion. SpeechLLM corresponds to the single-stage training strategy.}    \label{tab:data_expansion}
%\vspace{-6.5pt}
\end{table*}

\subsection{Fine-tuning a SpeechLLM Foundational Model}
 In this section, we highlight a major difference between the model we fine-tuned and a model that can be considered as a speechLLM foundational model. We considered building a task-specific system, that leverages the strengths of unimodal foundational models, but lack the training data scale to be categorized as a foundational model of its own. We consider our model as a \textit{composite} speechLLM of two unimodal foundational models but fine-tuned for a specific task in bimodal manner. 
 
 Among several open source models available, Qwen2-Audio model \cite{chu2024qwen2audiotechnicalreport}, can be considered a \textit{foundational} speechLLM, in contrast to our \textit{composite} model, having been trained on diverse and large datasets at scale in a multitask and multistage manner including pretraining, instruction-supervised training followed by DPO. We fine-tuned this model using QLORA applied to all its linear layers. 
 
 The results are presented in Table \ref{tab:qwen2_audio}. 
 The performance differences between the two models can be attributed to both the size of the models in terms of parameters, and the scale of data used for bimodal fine-tuning in addition to the training strategies employed.
 
 A larger model with extensive bimodal training is more likely to develop strong cross-modal representations enabling better generalization, as demonstrated by the results in Table~\ref{tab:qwen2_audio}. 
 
 A smaller model, by contrast, or one with limited bimodal fine-tuning, may rely more on unimodal strengths, leading to weaker multimodal alignment and requiring more task-specific adaptation or multitask adaptation by expanding the dataset, as demonstrated in Section~\ref{sec:data_expansion}.
 
 We also demonstrated the poor performance of the foundational model out-of-the-box, after extensive prompt engineering to steer the model for our specific task. This further illustrates the necessity of fine-tuning for specific tasks.
\begin{table*}[!b]
%\small
\centering
    \begin{tabular}{lccc} 
        \hline
        \textbf{System} & \textbf{CallCenter-A} &\textbf{CallCenter-B ID} & \textbf{CallCenter-B OOD} \\
                & Prec/Recall/F1    &  Prec/Recall/F1  & Prec/Recall/F1 \\
        \hline
        FT-LLM & 0.8523/ 0.9901/ 0.9160 & 0.8360/ 0.9530/ 0.8907  & 0.5439/ 0.8474/ 0.6626 \\
        FT-Qwen2-Audio & 0.7710/ 0.9799/ 0.8630  & 0.6708/ 0.8697/ 0.7574 &  0.4507/ 0.7619/ 0.5664 \\
        SpeechLLM-Multistage-C & 0.6842/ 0.9391/ 0.7916 & 0.4346/ 0.7588/ 0.5527 &  0.2653/ 0.4545/ 0.3351\\
        FT-(Whisper\textbar{}LLM) & 0.7006/ 0.8982/ 0.7872 & 0.5909/ 0.7067/ 0.6436 & 0.4280/ 0.6278/ 0.5090 \\
        SpeechLLM++ &  0.6632/ 0.9458 / 0.7797 & 0.4240/ 0.7810/ 0.5496 & 0.2645/ 0.5448/ 0.3561 \\
        SpeechLLM &  0.6553/ 0.9268/ 0.7677 & 0.3964/ 0.7306/ 0.5140 & 0.2386/ 0.4500/ 0.3119  \\
        PE-Qwen2-Audio & 0.1757/ 0.6334/ 0.2751  & 0.1059/ 0.6178/ 0.1807 & 0.0116/ 0.6034/ 0.0227 \\      \hline
    \end{tabular}
    \caption{Performances of prompt-engineered (PE) and fine-tuned (FT) Qwen2-Audio, with relevant results from earlier tables for better comparison. We extended earlier Multistage-C results to CallCenter-B dataset for completeness.}    \label{tab:qwen2_audio}
    %\vspace{8pt}
\end{table*}

\section{Conclusions}
In this paper, we have introduced speechLLMs for the open-ended slot-filling task and explored various architectural components, training strategies and multitask training to progressively improve performance over a simple baseline. 

Through our experiments we learned that adapter size plays a crucial role in model performance, but training difficulty scales with the adapter size and type, requiring careful selection. 

We also found that freezing the language model to preserve its original skills is not an effective strategy for tasks where the model lacks strong capabilities. Instead, aligning input representations with output generation by fine-tuning the LLM is critical for significantly improving task performance. 

To further enhance generalization and robustness we applied multistage training, which helped mitigate difficulties in single-stage joint learning, accelerated convergence, and improved learning performance. Additionally, expanding the dataset with diverse tasks, even if they were not the primary focus proved beneficial for overall performance. 

While these strategies helped to narrow the in-domain performance gap, surpassing the cascaded baseline system and approaching a text-based upper bound, challenges remain, particularly, in out-of-domain generalization and zero-shot capabilities. 

Also we highlighted trade-offs between \textit{composite} and \textit{foundational} models, emphasizing the need for continued scaling and refined training strategies to further improve robustness and generalization in real world scenarios. 

We believe this work provides practical insights and guidance for industry practitioners navigating the challenges of multimodal model development. Our findings highlight key tradeoffs in model design, training strategies, and data selection, offering valuable directions for those looking to build efficient, high-performing and adaptable multimodal systems for real-world use cases and computational constraints.

%\section{Conclusion}
%In this paper, we explored and demonstrated that fine-tuning for a specific task/domain, which has become an effective standard in the age of LLMs, yields similar satisfactory results for Speech LLMs. However, given a large number of possibilities and opportunities introduced by the architecture, special attention is required in selecting the components, hyper-parameters, training strategies, and data. Despite narrowing the gap, both robustness, generalization and zero-shot abilities still remain strong challenges. We believe that this work provides practical insights and guidance for industry practitioners navigating the challenges of multimodal model development. Our findings highlight key tradeoffs in model design, training strategies, and data selection, offering valuable directions for those looking to build efficient, high performing and adaptable multimodal systems within real world use cases and computational constraints. 

\clearpage
\section{Limitations}

During fine-tuning, for each configuration, the model with the best development loss is saved as the final model. Due to compute constraints, we reported only a single performance estimate per model configuration. We acknowledge the limitation of this approach; the lack of estimates of performance variance makes it difficult to assess the statistical significance of observed performance differences when they are relatively small. Therefore, it is important to view the reported numbers as representative rather than absolute. 

In addition, informed by earlier findings~\cite{rana-etal-2025-zero} that the results with/without human annotations can yield comparable outcomes, we chose not to rely on comprehensive human annotations or validations to keep cost low. Instead, we conducted limited validation to have an informal confirmation of quality using a relatively small subset of data generated by LLMs. 

Our study does not report results on open datasets, as no public benchmark exists at the scale we considered. Our dataset cannot be distributed, but reproducibility has been ensured through sufficient documentation for an experienced researcher to reproduce the major findings of this work. Specifically, the appendices include the full experimental configurations as well as the exact prompt templates used for data preparation and instruction creation.  Furthermore, the methodology and process are not tied to this specific dataset and can be applied to any similarly structured customer interaction data organized turn-by-turn.

Due to the variety of available text LLMs with varying scales, capabilities and training strategies along with the growing number of open source foundational speechLLMs, and driven by the paper length limit, we made a deliberate decision to present the results on a few representative models. We acknowledge that this may not cover the full diversity of the current model landscape. However, the inclusion of a larger foundational model, namely Qwen2-Audio, offers valuable insight into the scaling behavior of the composite model we considered. We believe this helps contextualize our results within the broader trajectory of model and data scaling. The results omitted here for brevity and clarity were in line with the observed trajectory.

%\section*{Acknowledgments}
\bibliography{paper}

%\clearpage
\appendix

\section{Evaluation Dataset Creation and Characteristics}
\label{sec:appendixA} 
To complement our primary in-domain evaluation dataset CallCenter-A, we introduce CallCenter-B dataset, sourced from different call-center conversations, to systematically assess the model's robustness, generalization and zero-shot abilities. CallCenter-B dataset differs from CallCenter-A in acoustic conditions, while partially overlapping in label distribution. Our analysis shows that 48\% of the slot labels in CallCenter-B  overlap with those in CallCenter-A dataset providing a subset that matches the linguistic characteristics of the training data. The remaining 52\% consist entirely of new labels not seen during training making it a strong test for zero-shot abilities. 
\begin{table*}[t]
    \centering
    \begin{tabular}{|l|c|}  
        \hline
        \textbf{Evaluation Dataset} & \textbf{\# samples}\\
        \hline
        CallCenter-A & 3253  \\
        CallCenter-B (ID) & 1482  \\
        CallCenter-B (OOD)& 3013  \\
        \hline
    \end{tabular}
    \caption{Evaluation Datasets: CallCenterA is in-domain for both speech and slots, CallCenter-B (ID) is in-domain for slots and out-of-domain for speech, and CallCenter-B (OOD) is out-of-domain for both slots and speech. }
    \label{tab:evaluation_datasets} 
\end{table*}
The sizes of evaluation sets used in our experiments are listed in Table~\ref{tab:evaluation_datasets}. Below we give additional information about partitions of these datasets used in experiments:

\begin{itemize}
    \item \textbf{Held-out test set}: CallCenter-A subset with the same acoustic and language characteristics as the training data, providing a benchmark for in-domain performance
    \item \textbf{Textually overlapping, acoustically  different set}: (CallCenter-B, ID) dataset containing language similar to the training data sharing the slots seen during the training, but with new acoustic conditions, testing the model's robustness to variations in speech input
    \item \textbf{Textually and acoustically unseen set}: (CallCenter-B, OOD) represents the same acoustic conditions as the second set, but with entirely unseen types of slots. This evaluates the model's ability to generalize to unseen language structures through novel slots, robustness to the shifted acoustic domain, and zero-shot performance.  
\end{itemize}
By comparing performance across CallCenter-A, CallCenter-B (ID) and (CallCenter-B (OOD) we gain deeper insights into the model's ability to handle real-world variability and its effectiveness in generalizing beyond the training distribution.   

\section{SIT and SQIT Datasets}
\label{sec:appendixA2}
The spoken instruction task (SIT) and spoken query–instruction task (SQIT) are derived from the text-based Alpaca dataset. The original Alpaca corpus contains two example formats:
(i) \{instruction, input, output\}, where instruction specifies the task, input provides context, and output is the expected response; and
(ii) \{instruction, output\}, where the instruction is self-contained and does not require additional context.

For SIT, speech is generated from the input field while the instruction remains text, allowing the model to process spoken input conditioned on a textual prompt. Here, the instructions are applied to spoken content. That is, a task is accomplished \textbf{on speech}.
For SQIT, speech is generated directly from the instruction field, with no accompanying text prompt, requiring the model to interpret spoken queries directly. Here, spoken content is used as instruction. That is, a task is accomplished \textbf{using speech}.

\section{Experimental Setup and Details}
\label{sec:appendixC}

\begin{table*}[t]
\centering
\setlength{\tabcolsep}{3pt} 
\begin{tabular}{|l|r|}
\hline
\textbf{Parameter} & \textbf{Value} \\
\hline
GPUs & 4 \\
GPU Memory & 24GB per GPU \\
LORA Rank & 32 \\
LORA Alpha & 128 \\
Dropout Rate & 0.05 \\
Batch Size per GPU & 4 \\
Gradient Accumulation Steps & 8 \\
Effective Batch Size & 128 \\
Maximum Learning Rate & 2e-4 \\
Number of Epochs & 10-15 \\
Warm-up & 20\% of iterations \\
AdamW beta1 & 0.9 \\
AdamW beta2 & 0.999 \\
AdamW epsilon & 1e-8 \\
Weight Decay & Not applied \\
Gradient Clipping Threshold & 1.0 \\
Adaptation Layers &  All linear layers \\
\hline
\end{tabular}
\caption{Fine-tuning hyperparameters and configuration}
    \label{tab:hyperparams} 
%\vspace{-10pt}
\end{table*}

Model development and experiments were conducted on hardware consisting of four NVIDIA A10G GPUs, each with 24GB of memory. 
The software ecosystem for this work was primarily based on the Hugging Face framework. Considering limited resource development and experimentation, to optimize the training/fine-tuning processes, we employed several advanced techniques. PEFT (parameter-efficient fine-tuning) was used to allow for fine-tuning of large models with significantly fewer parameters, reducing computational requirements. We also used QLORA (quantized low-rank adaptation), which uses 4-bit quantization. This technique drastically reduces the memory footprint of the model while maintaining performance. Additionally, we utilized Accelerate and DeepSpeed libraries to further optimize the training process.  For the LORA settings, we selected a rank of 32 with a corresponding $\alpha$ of 128. These settings control the complexity and capacity of the LoRA adaptation. We applied a dropout rate of 0.05 to all linear target modules to prevent overfitting. The batch size was set to 4 per GPU, with gradient accumulation of 8 steps, resulting in an effective batch size of 128. The learning rate was managed using a cosine scheduler over 10-15 epochs, with a maximum learning rate of $2 \times 10^{-4}$. We implemented a linear warm-up for the first 20\% of the total number of iterations. For optimization, we employed the AdamW (weighted Adam) optimizer. We used its default parameters, including $\beta_1 = 0.9$, $\beta_2= 0.999$, and $\epsilon = 10^{-8}$ , without applying any weight decay. We applied gradient clipping with a threshold of 1.0. 
During fine-tuning, for each configuration, the model with the best evaluation loss is saved as the final model. Due to compute constraints, we report point estimates of model performances in the experimental sections, while we acknowledge the limitations of this approach. We noticed that the final models may not correspond to the best performing models on the target metrics (precision/recall/F1). We observed performance fluctuation across runs and checkpoints with differences of $\pm 2$ percent points depending on the selection point. Therefore, it is important to review the reported numbers as representative rather than absolute. After the fine-tuning process, we implemented an efficient inference pipeline to evaluate the fine-tuned models. We used greedy search with temperature 0. The maximum number of new tokens was set to 512. This temperature setting without sampling is particularly useful when we want consistent, high-confidence responses from the model. The choice of  new tokens maximum length allows for reasonably lengthy responses while preventing excessively long generations. Table~\ref{tab:hyperparams} gives a summary of the hyperparameters and configuration settings used in fine-tuning, including hardware specifications, LoRA settings, optimization parameters, and training details.

\begin{figure}[h]
  \centering
  %\small
  \includegraphics[width=\linewidth]{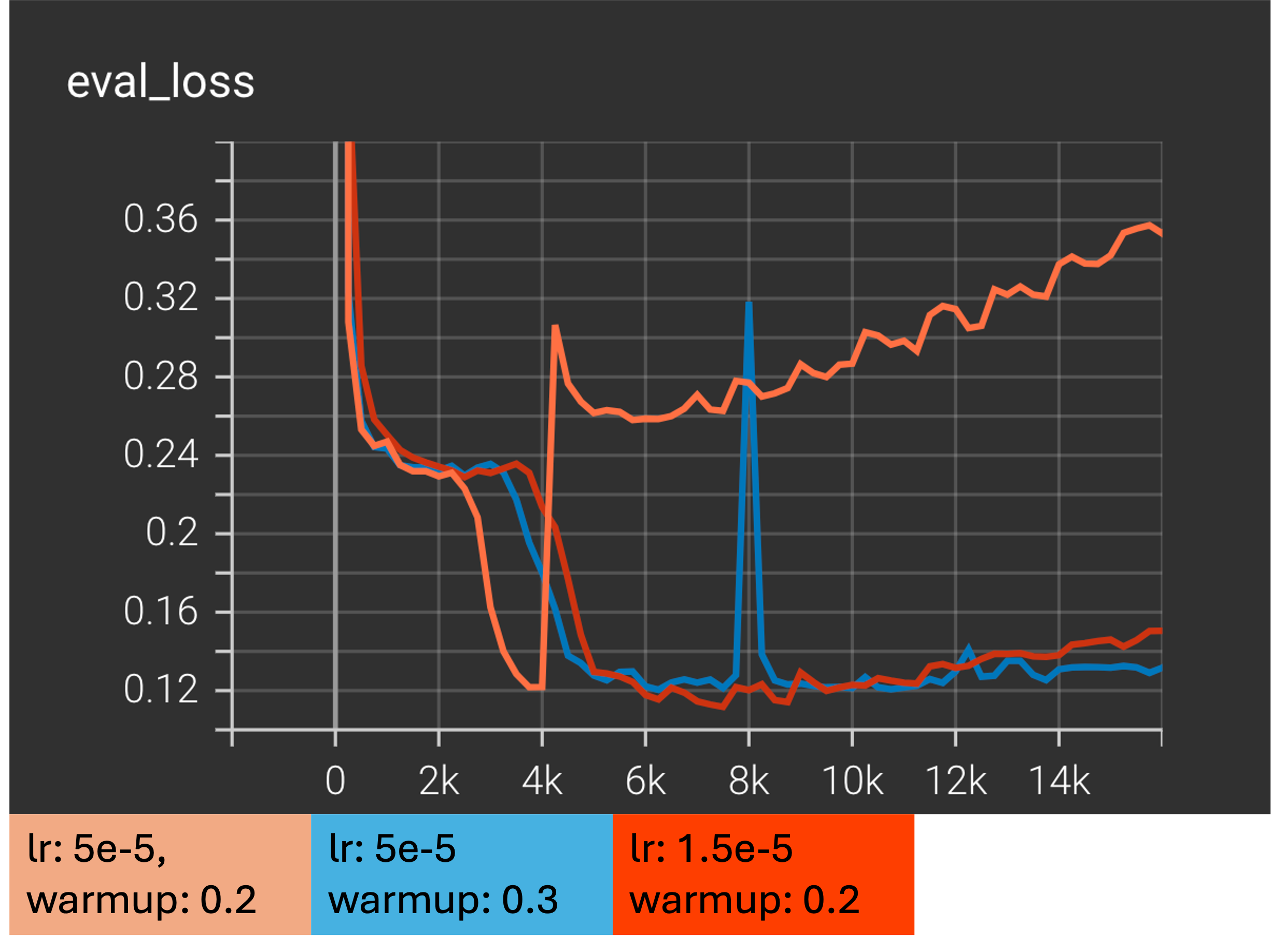}
  \caption{Learning curves showing training difficulties}
  \label{fig:learning_curves}
\end{figure}
\section{Training Difficulties}
\label{sec:appendixD}
In~\citet{das2024speechverselargescalegeneralizableaudio}, the authors observe frequent gradient explosion leading to suboptimal convergence when training both the adapter and LoRA from scratch. However, among the modules we experimented with as modality adapters, we observed that training instabilities first emerged when using the transformer. It is well-known that transformers are challenging to train due to issues such as vanishing/exploding gradients, sensitivity to learning rate scheduling, and difficulty in optimizing the attention architecture \citet{liu2023understandingdifficultytrainingtransformers}. Various tricks of the trade, such as using warm-up with proper learning rate scheduling, adding gradient clipping, applying  better initialization of weights, positioning of layer normalization and considering smaller architecture, are commonly used to stabilize training. Although all of these are considered in our training process, the default settings which have been largely robust for other adapter types, have turned out to be suboptimal for the transformer adapter. Training loss during the default warm-up period with the default learning rate has shown relatively large loss spikes. Based on the insights in \citet{kalra2024warmuplearningrateunderlying}, we experimented with reducing the maximum learning rate  or increasing the duration of the warm-up period. Our goal was to mitigate instability in regions where we observed
sharp training loss spikes, ensuring that learning continues with lower learning rates through those unstable phases. Reducing the learning rate worked better than increasing the warm-up duration considering the smoothness of the learning curve, but the final performance of the model was better with increased warm-up duration as indicated in Table~\ref{tab:adapter_lora}.  The learning curves in Figure \ref{fig:learning_curves} illustrate this divergent behavior with a relatively large learning rate, as well as stabilization by lowering the learning rate either by decreasing the maximum learning rate or increasing the warm-up duration.

\section{Learning Curves of Multi-Stage Training}
\label{sec:appendixE}
\begin{figure}[h]
  \centering
  %\small
  \includegraphics[width=\linewidth]{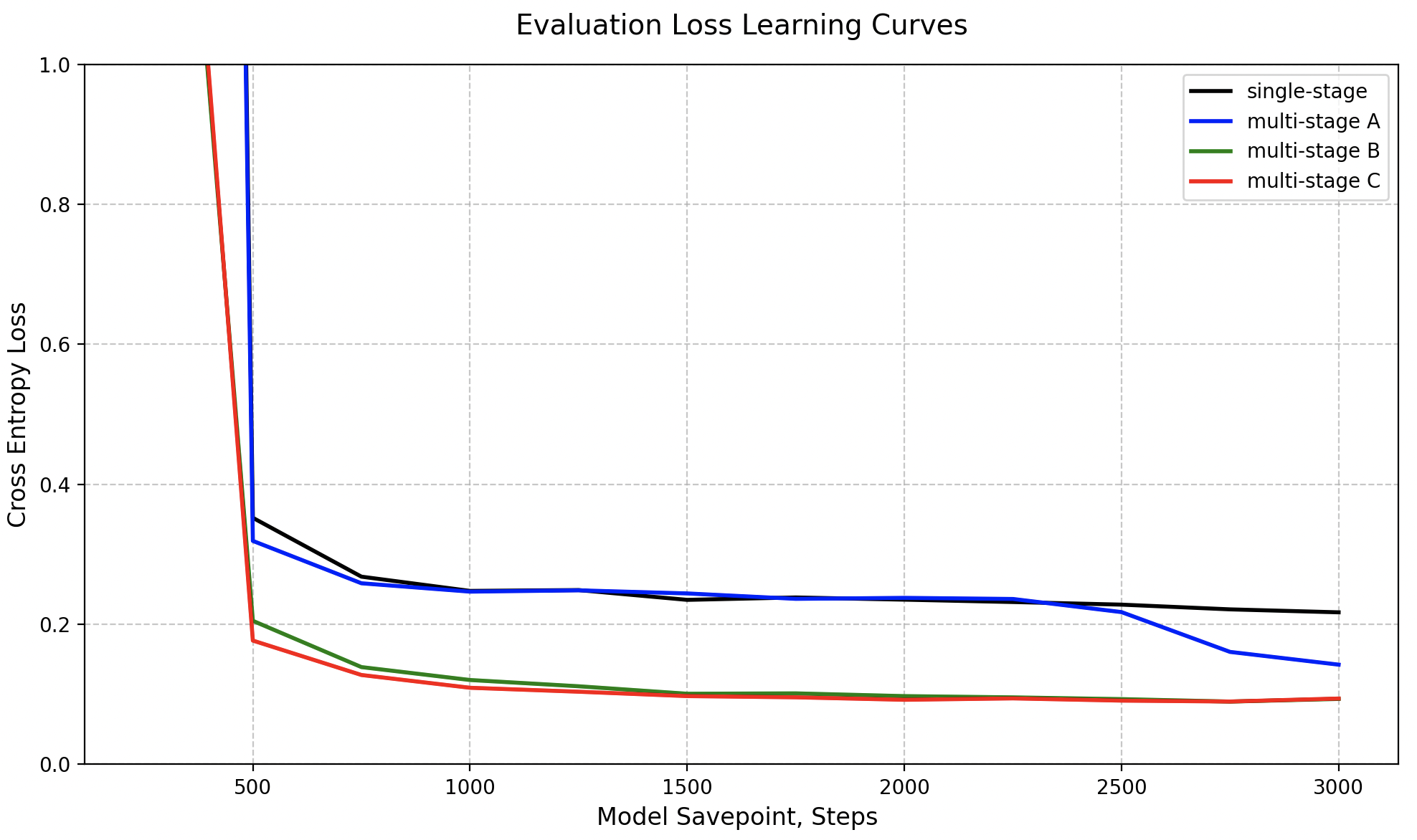}
  \caption{Learning curves comparing multistage training against single-stage joint fine-tuning.}
  \label{fig:single_vs_multi_stage}
\end{figure}

Figure~\ref{fig:single_vs_multi_stage} shows the cross-entropy loss over training checkpoints for four strategies discussed in the main text. The single-stage strategy represents the baseline, while the other strategies are variants of the multi-stage strategy. The curves demonstrate a clear progression in convergence speed and final performance, with each multi-stage strategy outperforming the baseline. The multi-stage C strategy achieves the fastest convergence and lowest loss, demonstrating the effectiveness of the multi-stage fine-tuning.  

\begin{figure*}[t]
\centering
\begin{minipage}{2.0\columnwidth}
\footnotesize
\centering
\fbox{%
\begin{tabular}{p{0.9\columnwidth}}
\textbf{[System]}: You are an expert in Natural Language Processing.

\textbf{[User]}: Your task is to identify all slots with their types and values in the given dialogue text between an agent and the customer turn by turn. The agent starts the dialogue. Return the output in a json format for every turn in the order given below between 'dialogue starts' and 'dialogue ends'  where key names  "normalized$\_$text" for a turn line and "slots" for  slot types and value as $\{$slot$\_$type: slot$\_$value$\}$ dictionary items. If there are no slot types in the line, return NA. For "normalized-text", DON'T change the content by rephrasing, auto-correcting, splitting, combining and skipping words. Punctuate by adding all the required punctuation marks, then capitalize appropriately, and then apply all the text normalization rules such as numbers to digits, currencies to symbols, dates and times to readable form of the given text. Slots will ONLY be for very SPECIFIC mentions of things in real world, entities, named entities, events by customer and agent. Avoid abstract, description like mentions. Slot values should be normalized too. Don't skip any line. Output only the JSON format\\
Dialogue starts:\\
$\{$dialog$\_$transcript$\}$\\
Dialogue ends:\\[0.5em]
\textbf{[Assistant]}: \\
```json\\
\hspace*{1em}[\\
\hspace*{2em}       $\{$
       
     \hspace*{3em}         "normalized$\_$text":  "Thank you for calling Wetsuits in Harrisburg, this is Heaven.",
        
        \hspace*{3em}        "slots": $\{$
        
            \hspace*{3em}    "location": "Harrisburg",
            
            \hspace*{3em}    "agent$\_$name": "Heaven"
        
          \hspace*{3em}    $\}$
    
  \hspace*{2em}    $\}$,
    
   \hspace*{2em} $\{$

    \hspace*{3em}    "normalized$\_$text": "Yeah, is Matt back yet on his trip?",
        
     \hspace*{3em}   "slots": $\{$
            
         \hspace*{3em}   "person$\_$name": "Matt"
        
       \hspace*{3em} $\}$
    
   \hspace*{2em} $\}$,

.........

.........

\hspace*{1em}]

```
\end{tabular}%
}
\caption{The prompt used for GPT-4o, [System]+[User] and its response/completion as [Assistant] which is the desired JSON annotation for turns in the dialog transcript. The $\{$dialog$\_$transcript$\}$ is the complete human transcription of the conversation. All the named entities mentioned in the examples are fictitious.}
\label{fig:gpt-prompt}
\end{minipage}
\end{figure*}
\section{Annotation and Task Prompt Examples}
\label{sec:appendixB}
The structure of the prompt for GPT-4o to annotate each turn in a complete conversation sample between an agent and a user is shown in Figure~\ref{fig:gpt-prompt}.  
In Figure~\ref{fig:task-prompts}, a subset of instructions used for the training data of the SpeechLLM model is shown.

\begin{figure*}[t]
\centering
\begin{minipage}{2.0\columnwidth}
\footnotesize
\centering
\fbox{%
\begin{tabular}{p{0.9\columnwidth}}
"Using the previous context, identify and extract slot values from $\{$slots$\}$ in the current utterance. Do not extract slot values from the context itself. Output should be in JSON format. Context: ``` $\{$context$\}$ ```",\\[0.5em]
"Based on the previous context, find slot values for $\{$slots$\}$ in the current speech. Ensure that slot extraction is only from the current utterance, not the context. Output in JSON. Previous context: ``` $\{$context$\}$ ```",\\[0.5em]
"Refer to the previous context to help identify slot values from $\{$slots$\}$ in the current audio. Slots should be extracted from the current utterance only. Output must be in JSON format. Context: ``` $\{$context$\}$ ```",\\[0.5em]
"Utilize the previous context to extract slot values from $\{$slots$\}$ in the current utterance. Do not consider the context for slot extraction. Format output as JSON. Context: ``` $\{$context$\}$ ```",\\[0.5em]
"Identify slot values for $\{$slots$\}$ in the current speech, using the previous context for guidance. Slot values should not be taken from the context itself. Output in JSON format. Context: ``` $\{$context$\}$ ```",
\end{tabular}%
}
\caption{A subset of prompts used for the ``instruction'' part of the training triplet $\{$audio, instruction, response$\}$. The fields ``slots'' and ``context'' in each prompt are populated as described in Section 2.2.}
\label{fig:task-prompts}
\end{minipage}
\end{figure*}

\end{document}